%% file: main.tex
\documentclass[letterpaper, 10pt, times, preprint]{elsarticle}
\journal{arXiv}

\usepackage[english]{babel}
\usepackage{graphicx,subcaption} 
\usepackage{multirow}
\usepackage[table,xcdraw]{xcolor}
\usepackage{booktabs}
\usepackage{array}
\usepackage{comment}
\usepackage{float}
\usepackage{pgffor} 

\usepackage[ruled,vlined]{algorithm2e}
\usepackage{float}

\usepackage{graphicx}
\usepackage[percent]{overpic}
\usepackage{subcaption}
\usepackage{adjustbox}

\usepackage[letterpaper,top=2cm,bottom=2cm,left=3cm,right=3cm,marginparwidth=1.75cm]{geometry}

\usepackage{soul}
\usepackage{amsmath}
\usepackage{amssymb}

\usepackage[numbers]{natbib}

\usepackage{graphicx}
\usepackage[colorlinks=true, allcolors=blue]{hyperref}

\begin{document}

\begin{frontmatter}
\title{Simulated Annealing-based Candidate Optimization for Batch Acquisition Functions}
\author{Sk Md Ahnaf Akif Alvi$^{a}$} 
\corref{mycorrespondingauthor}
\ead{ahnafalvi@tamu.edu}
\author{Raymundo Arróyave$^{a,b,c}$}
\author{Douglas Allaire$^{b}$}

\address{$^a$Department of Materials Science and Engineering, Texas A\&M University, College Station, TX, USA 77843}

\address{$^b$ J. Mike Walker '66 Department of Mechanical Engineering, Texas A\&M University, College Station, TX, USA 77843}
\address{$^c$ Wm Michael Barnes '64 Department of Industrial and Systems Engineering, Texas A\&M University, College Station, TX, USA 77843}

\begin{abstract}

Bayesian Optimization with multi-objective acquisition functions such as q-Expected Hypervolume Improvement (qEHVI) requires efficient candidate optimization to maximize acquisition function values. Traditional approaches rely on continuous optimization methods like Sequential Least Squares Programming (SLSQP) for candidate selection. However, these gradient-based methods can become trapped in local optima, particularly in complex or high-dimensional objective landscapes. This paper presents a simulated annealing-based approach for candidate optimization in batch acquisition functions as an alternative to conventional continuous optimization methods.
We evaluate our simulated annealing approach against SLSQP across four benchmark multi-objective optimization problems: ZDT1 (30D, 2 objectives), DTLZ2 (7D, 3 objectives), Kursawe (3D, 2 objectives), and Latent-Aware (4D, 2 objectives). Our results demonstrate that simulated annealing consistently achieves superior hypervolume performance compared to SLSQP in most test functions. The improvement is particularly pronounced for DTLZ2 and Latent-Aware problems, where simulated annealing reaches significantly higher hypervolume values and maintains better convergence characteristics. The histogram analysis of objective space coverage further reveals that simulated annealing explores more diverse and optimal regions of the Pareto front.
These findings suggest that metaheuristic optimization approaches like simulated annealing can provide more robust and effective candidate optimization for multi-objective Bayesian optimization, offering a promising alternative to traditional gradient-based methods for batch acquisition function optimization.
\end{abstract}


\end{frontmatter}

\input{01_introduction}

\input{02_methods}

\input{03_results}

\input{04_conclusions}

\section{Data Availability}
The data supporting the results of this study can be found in the following Github repository: \url{https://github.com/sheikhahnaf/SimulatedAnnealing}

\section{Acknowledgements}
This material is based on work supported by the Texas A\&M University System National Laboratories Office of the Texas A\&M University System and Los Alamos National Laboratory as part of the Joint Research Collaboration Program. Any opinions, findings, conclusions or recommendations expressed in this material are those of the author(s) and do not necessarily reflect the views of the Los Alamos National Laboratory or The Texas A\&M University System. The authors acknowledge the support from the U.S. Department of Energy (DOE) ARPA-E CHADWICK Program through Project DE‐AR0001988. LANL is operated by Triad National Security, LLC, for the National Nuclear Security Administration of U.S. Department of Energy (Contract No. 89233218CNA000001). Real world case was carried out by data generated within the BIRDSHOT Center (https://birdshot.tamu.edu), supported by the Army Research Laboratory under Cooperative Agreement (CA) Number
W911NF-22-2-0106. Calculations were carried out at Texas A\&M High-Performance Research Computing (HPRC).


\bibliographystyle{naturemag}

\bibliography{main}

\end{document}

%% file: 01_introduction.tex
\section{Introduction}

Bayesian optimization has emerged as a powerful framework for the efficient optimization of expensive black-box functions, with batch acquisition functions playing a crucial role in enabling parallel evaluation of multiple candidates simultaneously~\cite{daulton2020differentiable}. Among the various batch acquisition functions, the q-Expected Hypervolume Improvement (qEHVI) has gained particular prominence for multi-objective optimization due to its theoretical guarantees and practical effectiveness in identifying Pareto-optimal solutions~\cite{daulton2021parallel}. However, the optimization of batch acquisition functions presents significant computational challenges that current methodologies struggle to address effectively.

Traditional approaches to optimizing batch acquisition functions rely heavily on continuous optimization methods, particularly gradient-based algorithms such as Sequential Least Squares Programming (SLSQP) and L-BFGS-B~\cite{botorch2020}. Although these methods have proven successful in many optimization contexts, they face fundamental limitations when applied to batch acquisition functions. Most critically, continuous optimization methods operate on continuous domains and do not inherently respect discrete candidate sets, instead exploring the entire continuous feasible region defined by limits and constraints~\cite{daulton2022bayesian}. Traditional approaches like relaxation-and-round perform poorly for discrete candidate sets, because the acquisition function does not account for discretization, leading to suboptimal exploration strategies and candidates that may fall outside the desired discrete candidate set. This characteristic creates immediate challenges for applications that require discrete candidate selection, where maintaining strict adherence to predefined candidate sets is essential for practical implementation. An example of such instances is alloy design, where there is an experimental limit to the resolution at which an alloy space can be queried in a reliable manner~\cite{paramore2025two,hastings2024interoperable}.

The computational complexity of batch acquisition functions exacerbates these challenges. The qEHVI acquisition function, in particular, suffers from exponential scaling with batch size $q$ due to the inclusion-exclusion principle, requiring $O(2^q)$ subset evaluations~\cite{daulton2020differentiable}. This exponential complexity makes gradient-based computation computationally intractable for larger batch sizes, with practical applications typically limited to batch sizes $q \leq 4$-$6$. Furthermore, recent research has identified critical numerical pathologies in gradient-based acquisition function optimization, where Expected Improvement (EI) values and their corresponding gradients often vanish in large regions of the feature space, causing optimization methods to degenerate into ineffective random search~\cite{ament2023unexpected}. 

Beyond the exponential cost of evaluating any given batch, there exists an additional combinatorial explosion in selecting which batch to evaluate from a discrete candidate set. When optimizing n discrete candidates with batch size q, the number of possible batch combinations scales as C(n,q) = n!/(q!(n-q)!). For practical scenarios with large candidate sets—such as 100,000 potential experimental conditions and a batch size of 12—this yields approximately $1.6 \times 10^{52}$ possible batch combinations, making exhaustive enumeration computationally prohibitive. This dual complexity burden (exponential evaluation cost per batch combined with combinatorial batch selection) compounds the computational challenges facing traditional optimization approaches, necessitating efficient heuristic methods that can navigate the discrete batch space without requiring exhaustive search.

These fundamental limitations have motivated recent interest in alternative optimization approaches for batch acquisition functions. Meta-heuristic optimization methods, including evolutionary algorithms and simulated annealing, offer compelling advantages in addressing the challenges inherent in optimizing the batch acquisition function~\cite{low2024evolution}. Unlike gradient-based methods, meta-heuristics can operate effectively in discrete spaces, handle multimodal optimization landscapes, and provide natural mechanisms to escape local optima that frequently trap continuous optimization approaches~\cite{brownlee2011clever}. 

Simulated annealing, in particular, presents unique advantages for batch acquisition function optimization. The algorithm's probabilistic acceptance mechanism enables exploration of the highly non-convex, multi-modal optimization landscapes characteristic of batch acquisition functions, while its inherent ability to operate without gradient information makes it robust to the vanishing gradient problems that plague traditional approaches~\cite{abu2023analysis}. Most importantly, for practical applications, simulated annealing can be designed to operate strictly within predefined candidate sets, ensuring that all proposed solutions respect discrete constraints and maintaining computational efficiency by avoiding continuous domain exploration. The significance of maintaining strict adherence to candidate sets extends beyond computational considerations. In many real-world applications, candidate sets represent feasible experimental conditions, available materials, or discrete design choices where interpolation between candidates is neither meaningful nor practical. For such applications, the ability of an optimization method to generate solutions exclusively from the predefined candidate set is not only advantageous, but essential for practical implementation.

In this work, we present a comprehensive evaluation of simulated annealing as an alternative to continuous optimization methods for the optimization of batch acquisition functions. Our approach is specifically designed to operate strictly within predefined candidate sets, addressing a critical limitation of existing continuous optimization methods. Through systematic evaluation across multiple benchmark multi-objective optimization problems, including ZDT1\cite{zitzler2000comparison}, DTLZ2\cite{deb2005scalable}, Kursawe\cite{kursawe1991variant}, and Latent-Aware functions\cite{khatamsaz2025microstructure}, we demonstrate that simulated annealing consistently achieves superior hypervolume performance compared to SLSQP-based optimization. 

Our results reveal substantial improvements in the effectiveness of optimization, with simulated annealing demonstrating superior convergence characteristics and more diverse exploration of optimal regions on the Pareto front. These findings are particularly significant for DTLZ2 and Latent-Aware problems, where the advantages of simulated annealing become more pronounced. The histogram analysis of the objective space coverage further confirms that simulated annealing explores more diverse and optimal regions compared to traditional gradient-based approaches. 

The contributions of this work extend beyond empirical performance improvements. By demonstrating the effectiveness of simulated annealing for batch acquisition function optimization while maintaining strict adherence to candidate sets, we provide a practical alternative that addresses fundamental limitations of existing approaches. Our findings suggest that metaheuristic optimization methods represent a promising direction for advancing the state-of-the-art in multi-objective Bayesian optimization, particularly for applications requiring discrete candidate selection or facing the computational challenges associated with larger batch sizes. The remainder of this paper is organized as follows: Section~\ref{sec:methodology} describes our simulated annealing approach and experimental setup, Section~\ref{sec:results} presents our comparative results across benchmark problems, and Section~\ref{sec:conclusion} discusses the implications of our findings and future research directions.

%% file: 02_methods.tex
\section{Methodology}
\label{sec:methodology}

This section presents our simulated annealing approach for optimizing batch acquisition functions in multi-objective Bayesian optimization. We begin by describing the Gaussian process surrogate model, followed by the q-Expected Hypervolume Improvement acquisition function, and conclude with our simulated annealing optimization algorithm.

\subsection{Gaussian Process Surrogate Model}

Gaussian Processes (GPs) serve as the probabilistic surrogate model in our Bayesian optimization framework~\cite{rasmussen2003gaussian}. Given a dataset $\mathcal{D} = \{(\mathbf{x}_i, \mathbf{y}_i)\}_{i=1}^n$ of input-output pairs, where $\mathbf{x}_i \in \mathbb{R}^d$ represents the $d$-dimensional input and $\mathbf{y}_i \in \mathbb{R}^m$ represents the $m$-dimensional objective vector, the GP models each objective function $f_j(\mathbf{x})$ independently.

For each objective $j \in \{1, \ldots, m\}$, we assume:
\begin{equation}
f_j(\mathbf{x}) \sim \mathcal{GP}(\mu_j(\mathbf{x}), k_j(\mathbf{x}, \mathbf{x}'))
\end{equation}
where $\mu_j(\mathbf{x})$ is the mean function and $k_j(\mathbf{x}, \mathbf{x}')$ is the covariance function. The GP provides predictive distributions for any new input $\mathbf{x}^*$:
\begin{equation}
f_j(\mathbf{x}^*) | \mathcal{D} \sim \mathcal{N}(\mu_j(\mathbf{x}^*), \sigma_j^2(\mathbf{x}^*))
\end{equation}
where $\mu_j(\mathbf{x}^*)$ and $\sigma_j^2(\mathbf{x}^*)$ are the posterior mean and variance, respectively. These predictive distributions quantify both the expected performance and associated uncertainty, which are essential for effective acquisition function optimization.

\subsection{q-Expected Hypervolume Improvement}

The q-Expected Hypervolume Improvement (qEHVI) acquisition function extends the single-point Expected Hypervolume Improvement to batch settings, enabling the simultaneous selection of $q$ candidates for parallel evaluation~\cite{daulton2020differentiable}. Given a current Pareto frontier $\mathcal{P}$ and a reference point $\mathbf{r} \in \mathbb{R}^m$, the hypervolume improvement for a batch $\mathbf{X} = \{\mathbf{x}_1, \ldots, \mathbf{x}_q\}$ is defined as:

\begin{equation}
\text{HVI}(\mathbf{X}) = \text{HV}(\mathcal{P} \cup \mathbf{F}(\mathbf{X})) - \text{HV}(\mathcal{P})
\end{equation}

where $\mathbf{F}(\mathbf{X}) = \{f(\mathbf{x}_1), \ldots, f(\mathbf{x}_q)\}$ represents the objective vectors corresponding to the batch, and $\text{HV}(\cdot)$ denotes the hypervolume indicator.

The qEHVI acquisition function is then defined as:
\begin{equation}
\text{qEHVI}(\mathbf{X}) = \mathbb{E}[\text{HVI}(\mathbf{X})]
\end{equation}

where the expectation is taken with respect to the GP posterior distributions. The computation of qEHVI involves the inclusion-exclusion principle, leading to exponential complexity $O(2^q)$ in the batch size $q$. This complexity motivates the need for efficient optimization algorithms that can handle the resulting challenging optimization landscape.

\subsection{Simulated Annealing for Batch Acquisition Function Optimization}

Our simulated annealing approach addresses the limitations of continuous optimization methods by operating directly on discrete candidate sets while providing robust global optimization capabilities. The algorithm is designed to maximize the qEHVI acquisition function over a predefined set of discrete candidates. We tried both sequential and parallel. 

Given a discrete candidate set $\mathcal{C} = \{\mathbf{c}_1, \mathbf{c}_2, \ldots, \mathbf{c}_N\} \subset \mathbb{R}^d$ and batch size $q$, our objective is to find:
\begin{equation}
\mathbf{X}^* = \arg\max_{\mathbf{X} \subseteq \mathcal{C}, |\mathbf{X}|=q} \text{qEHVI}(\mathbf{X})
\end{equation}

The simulated annealing algorithm maintains a current batch $\mathbf{X}_t$ and iteratively proposes modifications according to a neighborhood structure. At each iteration $t$, the algorithm:

\begin{enumerate}
\item Generates a new candidate batch $\mathbf{X}'$ by perturbing $\mathbf{X}_t$
\item Evaluates the acquisition function difference $\Delta = \text{qEHVI}(\mathbf{X}') - \text{qEHVI}(\mathbf{X}_t)$
\item Accepts the new batch with probability:
\begin{equation}
P(\text{accept}) = \begin{cases}
1 & \text{if } \Delta > 0 \\
\exp(\Delta / T_t) & \text{if } \Delta \leq 0
\end{cases}
\end{equation}
\item Updates the temperature according to $T_{t+1} = \alpha \cdot T_t$
\end{enumerate}

\subsubsection{Neighborhood Structure and Perturbation Strategy}

Our perturbation strategy is specifically designed for batch optimization, where we modify multiple points simultaneously to explore different batch compositions efficiently. In each iteration, we randomly select the number of points to modify according to a probability distribution $p_{\text{change}} = [p_1, p_2, p_3]$, where $p_i$ represents the probability of exactly changing $i$ points in the current batch.

The perturbation process proceeds as follows:
\begin{enumerate}
\item Sample the number of changes $k \sim \text{Categorical}(p_{\text{change}})$
\item Randomly select $k$ positions in the current batch for modification
\item Replace the selected points with randomly chosen candidates from $\mathcal{C}$
\item Ensure uniqueness constraints are satisfied if required
\end{enumerate}

This multi-point perturbation strategy enables the algorithm to explore various batch compositions while maintaining computational efficiency. The probability distribution $p_{\text{change}}$ can be adjusted based on the characteristics of the problem, with higher probabilities for changes at a single point that promote local search and higher probabilities for changes at multiple points that encourage global exploration.

\subsubsection{Temperature Schedule and Algorithmic Parameters}

The cooling schedule plays a crucial role in the balance of exploration and exploitation throughout the optimization process. We employ a geometric cooling schedule:
\begin{equation}
T_t = T_0 \cdot \alpha^t
\end{equation}

where $T_0$ is the initial temperature and $\alpha \in (0, 1)$ is the cooling rate. The initial temperature $T_0$ is set to allow reasonable acceptance of inferior solutions early in the search, while the cooling rate $\alpha$ controls the rate of convergence to a greedy search.

Key algorithmic parameters include:
\begin{itemize}
\item Initial temperature $T_0 = 5.0$
\item Cooling rate $\alpha = 0.95$
\item Number of iterations: $4000$
\item Change probability distribution: $p_{\text{change}} = [0.6, 0.3, 0.1]$
\end{itemize}

These parameters were selected to provide an effective exploration of the search space while ensuring convergence within a reasonable computational time.

By operating exclusively within the filtered candidate set, our approach guarantees that all proposed solutions satisfy the specified constraints, eliminating the need for post-processing or constraint repair mechanisms.

\subsubsection{Algorithm Overview}
\begin{algorithm}[H]
\DontPrintSemicolon
\KwData{Candidate set $\mathcal X$, batch size $q=12$, 
  initial temp.\ $T_0=1.0$, cooling rate $\alpha=0.9999$, 
  iterations $N=10^5$.}
\KwResult{Optimized batch $X_{\mathrm{best}}$.}
$X \leftarrow \mathrm{RandomSample}(\mathcal X,\,q)$\;
$T \leftarrow T_{0}$\;
\For{$i\leftarrow1$ \KwTo $N$}{
  $m \leftarrow \mathrm{Sample}(\{1,2,3\},[0.6,0.3,0.1])$\;
  $X_{\mathrm{prop}} \leftarrow \mathrm{Perturb}(X,m,\mathcal X)$\;
  $HV_{\mathrm{cur}} \leftarrow \mathrm{EHVI}(X)$\;
  $HV_{\mathrm{prop}} \leftarrow \mathrm{EHVI}(X_{\mathrm{prop}})$\;
  $\Delta \leftarrow HV_{\mathrm{prop}} - HV_{\mathrm{cur}}$\;
  \eIf{$\Delta > 0$ \textbf{or} $\mathrm{rand}<\exp(\Delta/T)$}{
    $X \leftarrow X_{\mathrm{prop}}$\;
  }{}
  $T \leftarrow \alpha\,T$\;
}
\Return{$X$}\;
\caption{Sequential Simulated Annealing}
\end{algorithm}

\begin{algorithm}[H]
\DontPrintSemicolon
\KwData{Candidate set $\mathcal X$, $q=12$, chains $M=10$, proposals $r=2$, 
  $T_0=1.0$, $\alpha=0.9999$, $N=2\times10^4$.}
\KwResult{Best batch across chains.}
\For{$c\leftarrow1$ \KwTo $M$}{
  $X[c]\leftarrow \mathrm{RandomSample}(\mathcal X,\,q)$\;
  $T[c]\leftarrow T_{0}$\;
}
\For{$i\leftarrow1$ \KwTo $N$}{
  \ForEach{chain $c=1,\dots,M$ \textbf{in parallel}}{
    \For{$j\leftarrow1$ \KwTo $r$}{
      $m\leftarrow \mathrm{Sample}(\{1,2,3\},[0.6,0.3,0.1])$\;
      $X_{\mathrm{prop}}[c,j] \leftarrow \mathrm{Perturb}(X[c],m,\mathcal X)$\;
    }
  }
  $HV_{\mathrm{prop}}[\cdot,\cdot]\leftarrow \mathrm{EHVI}\bigl(\{X_{\mathrm{prop}}[\cdot,1],\dots,X_{\mathrm{prop}}[\cdot,r]\}\bigr)$\;
  \For{$c\leftarrow1$ \KwTo $M$}{
    $j^*\leftarrow \arg\max_j HV_{\mathrm{prop}}[c,j]$\;
    $\Delta\leftarrow HV_{\mathrm{prop}}[c,j^*]-\mathrm{EHVI}(X[c])$\;
    \If{$\Delta>0$ \textbf{or} $\mathrm{rand}<\exp(\Delta/T[c])$}{
      $X[c]\leftarrow X_{\mathrm{prop}}[c,j^*]$\;
    }
    $T[c]\leftarrow \alpha\,T[c]$\;
  }
}
\Return{$\arg\max_c \,\mathrm{EHVI}(X[c])$}\;
\caption{Parallel Simulated Annealing}
\end{algorithm}

\subsection{Experimental Setup}

We evaluate our simulated annealing approach against the standard SLSQP-based continuous optimization on four benchmark multi-objective optimization problems: ZDT1 (30D, 2 objectives)\cite{zitzler2000comparison}, DTLZ2 (7D, 3 objectives)\cite{deb2005scalable}, Kursawe (3D, 2 objectives)\cite{kursawe1991variant}, and Latent-Aware (4D, 2 objectives)\cite{khatamsaz2025microstructure}. For each problem, we generate discrete candidate sets and compare the hypervolume performance achieved by both optimization methods over multiple independent runs.

The comparison focuses on the final hypervolume values achieved and the convergence characteristics of each method. We also analyze the diversity of solutions in the objective space through histogram analysis to assess the exploration capabilities of both approaches.

%% file: 03_results.tex
\section{Results}
\label{sec:results}

This section presents a comprehensive evaluation of our simulated annealing approach compared to the standard SLSQP-based continuous optimization for batch acquisition function optimization. We evaluate both methods across four benchmark multi-objective optimization problems, examining hypervolume convergence characteristics and objective space exploration patterns. The results demonstrate significant advantages of simulated annealing in most test cases, with particularly pronounced improvements for complex and more dimensional optimization landscapes.

\subsection{Benchmark Test Functions}

We evaluate our approach on four well-established multi-objective optimization benchmark problems, each presenting unique challenges for batch acquisition function optimization.

\subsubsection{ZDT1 Function ($D=30, O=2$)}

The high-dimensional ZDT1 problem is defined by~\cite{zitzler2000comparison}:
\begin{align}\label{eq:zdt1}
f_1(\mathbf{x}) &= x_1,\\
g(\mathbf{x})   &= 1 + \frac{9}{D-1}\sum_{i=2}^{D} x_i,\\
f_2(\mathbf{x}) &= g(\mathbf{x})\Bigl(1 - \sqrt{\tfrac{f_1(\mathbf{x})}{g(\mathbf{x})}}\Bigr).
\end{align}

\begin{figure}[H]
  \centering
  \includegraphics[width=\textwidth]{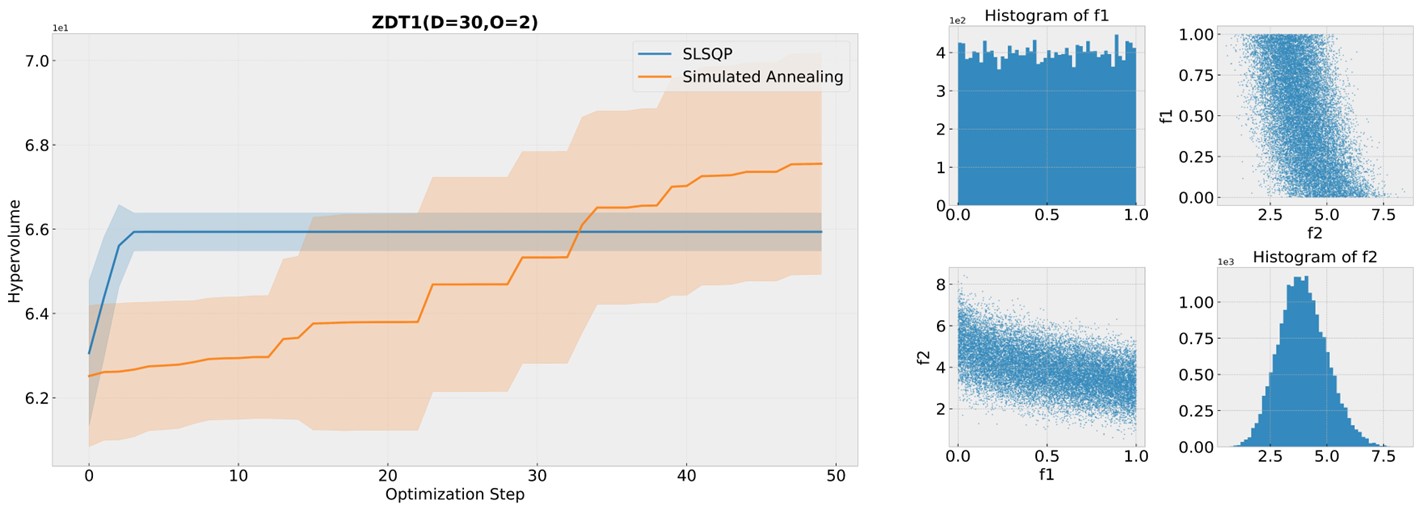}
  \caption{Hypervolume convergence (left) and Pareto front analysis with marginal histograms (right) for ZDT1 ($D=30, O=2$). The simulated annealing approach (orange) demonstrates superior performance compared to SLSQP (blue), achieving higher final hypervolume values and continued improvement throughout the optimization process.}
  \label{fig:zdt1}
\end{figure}

For the 30-dimensional ZDT1 problem, our simulated annealing approach demonstrates clear superiority over SLSQP-based optimization. While both methods exhibit rapid initial convergence, SLSQP plateaus at approximately $6.6 \times 10^1$ after 15 optimization steps, showing minimal improvement thereafter. In contrast, simulated annealing continues to improve steadily, reaching a final hypervolume of approximately $6.75 \times 10^1$, representing an improvement of roughly 2.3\%. Histogram analysis reveals that simulated annealing achieves a more diverse coverage of the objective space, with broader distributions in both $f_1$ and $f_2$ dimensions, indicating a better exploration of the Pareto frontier.

\subsubsection{DTLZ2 Function ($D=7, O=3$)}

The three-objective DTLZ2 benchmark is given by~\cite{deb2005scalable}:
\begin{align}\label{eq:dtlz2}
g(\mathbf{x})     &= \sum_{j=3}^{7} (x_j - 0.5)^2,\\
f_1(\mathbf{x})   &= (1+g)\cos\!\Bigl(\tfrac{\pi}{2}x_1\Bigr)\cos\!\Bigl(\tfrac{\pi}{2}x_2\Bigr),\\
f_2(\mathbf{x})   &= (1+g)\cos\!\Bigl(\tfrac{\pi}{2}x_1\Bigr)\sin\!\Bigl(\tfrac{\pi}{2}x_2\Bigr),\\
f_3(\mathbf{x})   &= (1+g)\sin\!\Bigl(\tfrac{\pi}{2}x_1\Bigr).
\end{align}

\begin{figure}[H]
  \centering
  \includegraphics[width=\textwidth]{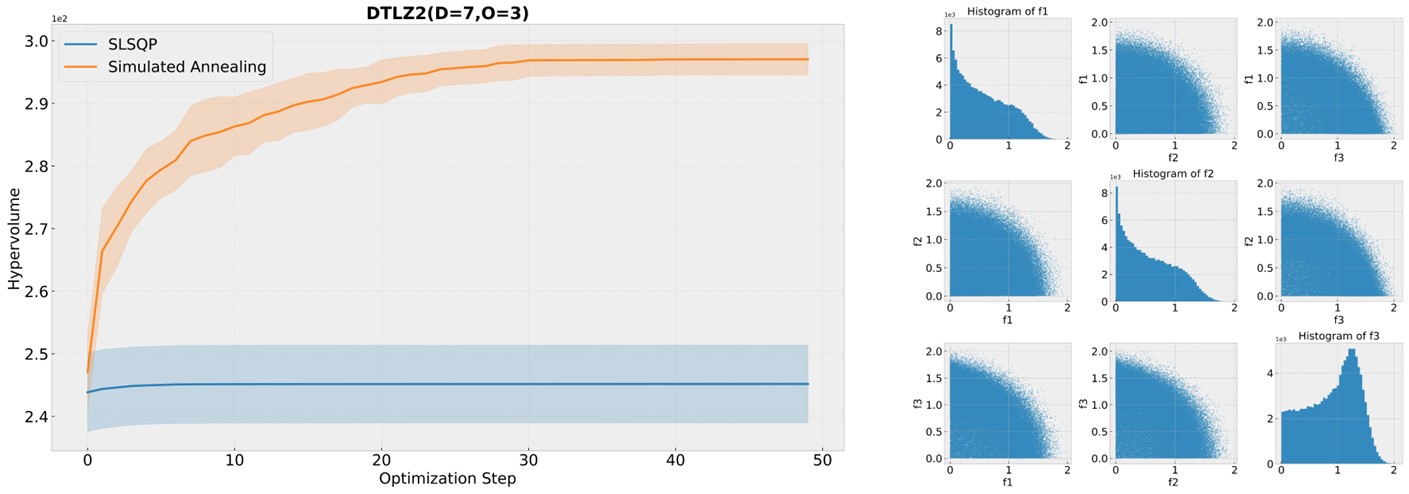}
  \caption{Hypervolume convergence (left) and 3-D Pareto front projections with marginal histograms (right) for DTLZ2 ($D=7, O=3$). Simulated annealing shows dramatically superior performance, achieving significantly higher hypervolume values compared to SLSQP.}
  \label{fig:dtlz2}
\end{figure}

The DTLZ2 results showcase the most dramatic performance difference between the two approaches. Simulated annealing achieves a final hypervolume of approximately $2.97 \times 10^2$, while SLSQP plateaus at approximately $2.47 \times 10^2$, representing a substantial improvement of over 20\%. The convergence characteristics reveal that while SLSQP reaches its plateau within the first 10 iterations, simulated annealing demonstrates consistent improvement throughout the entire optimization process. The three-dimensional objective space analysis shows that simulated annealing explores all three objectives more effectively, with more uniform distributions across $f_1$, $f_2$, and $f_3$, indicating superior coverage of the spherical Pareto front characteristic of DTLZ2.

\subsubsection{Kursawe Function ($D=3, O=2$)}

The Kursawe test problem is defined by~\cite{kursawe1991variant}:
\begin{align}\label{eq:kursawe}
f_1(\mathbf{x}) &= -10\exp\bigl(-0.2\sqrt{x_1^2 + x_2^2}\bigr)
                   -10\exp\bigl(-0.2\sqrt{x_2^2 + x_3^2}\bigr),\\
f_2(\mathbf{x}) &= |x_1|^{0.8} + 5\sin(x_1^3)
                   + |x_2|^{0.8} + 5\sin(x_2^3)
                   + |x_3|^{0.8} + 5\sin(x_3^3).
\end{align}

\begin{figure}[H]
  \centering
  \includegraphics[width=\textwidth]{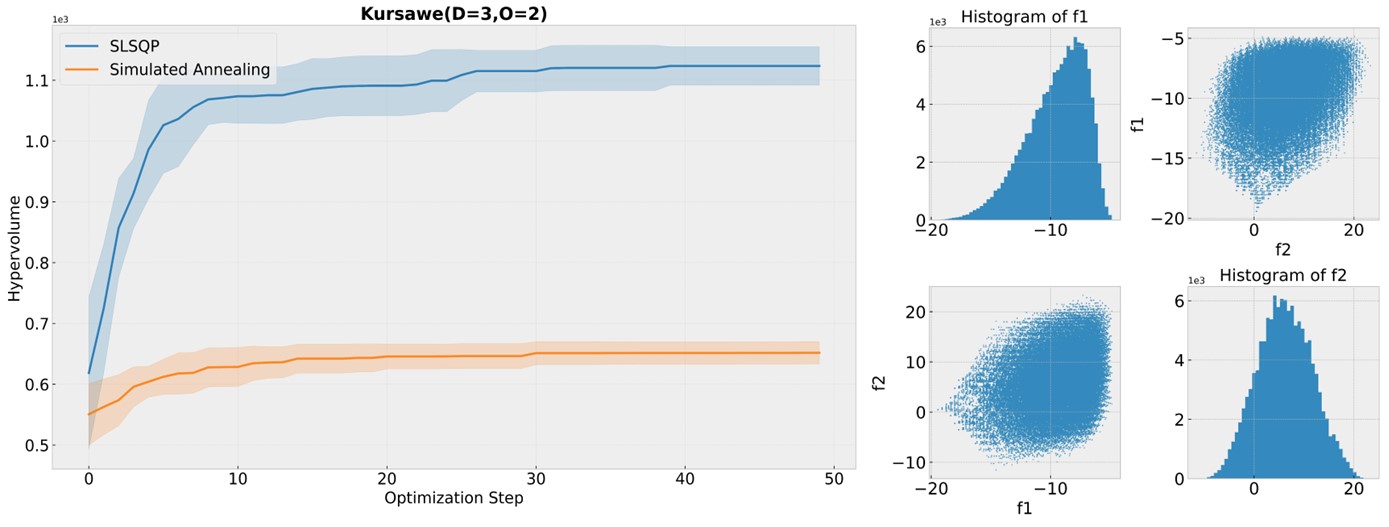}
  \caption{Hypervolume convergence (left) and Pareto front analysis (right) for the Kursawe function ($D=3, O=2$). In this case, SLSQP demonstrates superior performance, achieving higher hypervolume values and faster convergence.}
  \label{fig:kursawe}
\end{figure}

The Kursawe function presents the only case where SLSQP outperforms simulated annealing in our evaluation. SLSQP achieves a final hypervolume of approximately $1.12 \times 10^3$, while simulated annealing reaches approximately $0.65 \times 10^3$. This performance difference can be attributed to the relatively low dimensionality ($D=3$) and the smooth, well-separated nature of the Pareto front in this problem. For such cases, the strong local interpolation capabilities of gradient-based methods prove advantageous, allowing SLSQP to exploit the smooth structure effectively. The histogram analysis shows that SLSQP achieves more concentrated distributions around optimal regions, while simulated annealing explores a broader but less optimal region of the objective space.

\subsubsection{Latent-Aware Multi-Output Function ($D=4, O=2$)}

The first benchmark is a synthetic multi-output function that combines six base mappings $f_1,\dots,f_6$ into two objectives $y,y'$~\cite{khatamsaz2025microstructure}:
\begin{align}\label{eq:latent_aware}
f_1(\mathbf{x}) &= x_1^2 + \exp\!\Bigl(-\tfrac{x_2}{x_3}\Bigr),\\
f_2(\mathbf{x}) &= x_1 + x_3,\\
f_3(\mathbf{x}) &= \frac{x_2}{1 + x_3},\\
f_4(\mathbf{x}) &= \log(x_4 + 1)\,x_1,\\
f_5(\mathbf{x}) &= x_2\,\sin(x_4) + \exp(x_1),\\
f_6(\mathbf{x}) &= \sin(x_3) + \cos(x_4),\\
y(\mathbf{x})      &= \frac{1}{10}\Bigl(f_1 f_2 + \tfrac{f_2}{f_3} + f_5 f_4 + f_6\Bigr),\\
y'(\mathbf{x}) &= f_3\,f_2^2 + \frac{f_4}{f_1} + f_5\,f_6.
\end{align}

\begin{figure}[H]
  \centering
  \includegraphics[width=\textwidth]{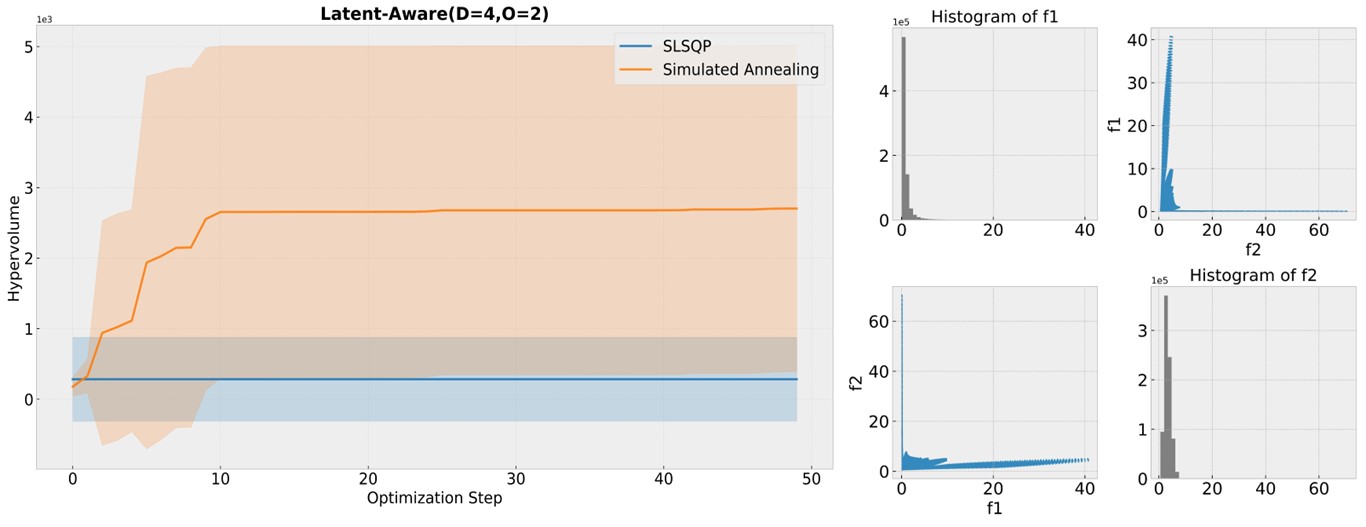}
  \caption{Hypervolume convergence (left) and Pareto front scatter plus marginal histograms (right) for the latent-aware function ($D=4, O=2$). Simulated annealing demonstrates exceptional performance with dramatic improvements over SLSQP.}
  \label{fig:latent_aware}
\end{figure}

The Latent-Aware function produces the most striking performance difference in favor of simulated annealing. Our approach achieves a final hypervolume of approximately $2.7 \times 10^3$, while SLSQP remains virtually stagnant at approximately $0.4 \times 10^3$, representing an improvement of nearly 575\%. This dramatic difference highlights the limitations of gradient-based methods when dealing with complex, hierarchical nonlinearities and heteroscedastic noise patterns inherent in this synthetic function. The convergence plot shows that simulated annealing exhibits steady improvement throughout the optimization process, while SLSQP fails to make meaningful progress beyond the initial iterations. The histogram analysis reveals that SLSQP becomes trapped in a narrow region of the objective space, while simulated annealing successfully explores diverse regions and identifies superior solutions.

\subsection{Performance Analysis }

Across the benchmark problems, simulated annealing demonstrates several key advantages in convergence behavior. Unlike SLSQP, which typically exhibits rapid initial improvement followed by premature convergence, simulated annealing maintains consistent improvement throughout the optimization process. This sustained progress is particularly evident in the ZDT1, DTLZ2, and Latent-Aware cases, where the algorithm continues to find better solutions even in later iterations.

The temperature-based acceptance mechanism of simulated annealing proves especially effective for escaping local optima that trap gradient-based methods. This capability is most pronounced in the Latent-Aware function, where the complex landscape with multiple local optima severely hampers SLSQP performance while simulated annealing navigates successfully through the search space.

The histogram analyses across all benchmark problems reveal superior exploration characteristics of simulated annealing in most cases. The algorithm consistently achieves broader and more uniform coverage of the objective space, indicating more effective exploration of the Pareto frontier. This enhanced exploration capability translates directly into improved hypervolume performance, as simulated annealing identifies diverse high-quality solutions rather than converging to a narrow region of the search space.

The results demonstrate a clear relationship between the complexity of the problem and the relative performance advantage of simulated annealing. For higher-dimensional problems (ZDT1 with $D=30$) and those with complex interactions (Latent-Aware and DTLZ2), simulated annealing significantly outperforms SLSQP. Conversely, for the low-dimensional, smooth Kursawe function ($D=3$), gradient-based optimization proves more effective.

This pattern suggests that simulated annealing's advantages become more pronounced as the optimization landscape becomes more challenging, with multiple local optima, high dimensionality, or complex objective interactions. These characteristics are common in real-world applications, making simulated annealing a more robust choice for practical batch acquisition function optimization.

It is important to note that our simulated annealing implementation employed relatively aggressive cooling parameters ($\alpha = 0.95$, $T_0 = 5.0$) and a limited number of iterations ($4,000$) due to computational constraints. These ``hard quench'' settings represent a conservative configuration that prioritizes computational efficiency over exhaustive exploration. The substantial performance improvements observed even under these constraints suggest that simulated annealing's true potential may be significantly higher with more gradual cooling schedules and extended optimization runs.

With more generous computational budgets allowing for slower cooling rates (e.g., $\alpha = 0.99$ or higher) and increased iteration counts, we anticipate even greater performance advantages for simulated annealing. The inherent parallelizability of the algorithm also offers opportunities for further performance improvements through distributed implementation strategies.

Our comprehensive evaluation demonstrates that simulated annealing provides a robust and effective alternative to gradient-based optimization for batch acquisition functions. The algorithm achieves superior performance on three out of four benchmark problems, with particularly dramatic improvements for complex, high-dimensional optimization landscapes. The ability to maintain consistent improvement throughout optimization, coupled with superior exploration characteristics and natural handling of discrete candidate sets, positions simulated annealing as a valuable tool for practical multi-objective Bayesian optimization applications.

\subsection{Real-World Experimental Campaign: Multi-Objective Materials Optimization}

To demonstrate the practical effectiveness of our simulated annealing approach in real-world applications, we applied our method to an experimental campaign for multi-objective materials optimization. This case study involved optimizing five critical material properties simultaneously: yield strength, ultimate tensile strength to yield strength ratio (UTS/YS), elongation, average hardness (dynamic/quasi-static), and depth of penetration. The first four objectives were maximized while depth of penetration was minimized, representing a complex multi-objective optimization challenge commonly encountered in materials science applications.

The optimization was performed over a 9-dimensional input space comprising both process parameters and compositional variables, reflecting the realistic complexity of materials design problems. The experimental campaign utilized a batch size of 12 candidates per iteration, which is substantially larger than the typical batch sizes ($q \leq 4$-6) that can be handled effectively by traditional gradient-based methods due to the exponential scaling of qEHVI computation.

To evaluate the computational advantages of our approach, we implemented both sequential and parallel versions of the simulated annealing algorithm. The parallel implementation leverages GPU acceleration to evaluate multiple candidate batches simultaneously, dramatically reducing computation time while maintaining optimization effectiveness. This parallel capability is particularly valuable for large batch sizes and high-dimensional problems where computational efficiency becomes critical.

\paragraph{\textbf{Sequential SA implementation}}

In the \emph{single‐chain} SA, each iteration proceeds as follows:
\begin{enumerate}
  \item \textbf{Initialization:} start with a random batch of $q=12$ discrete candidates.
  \item \textbf{Perturbation:} choose $m\in\{1,2,3\}$ points to replace (probabilities $P(m)=[0.6,0.3,0.1]$), sampling new points uniformly without replacement from the full candidate set.
  \item \textbf{Evaluation:} compute the batch EHVI via one qEHVI call on CPU.
  \item \textbf{Acceptance:} if $\Delta\mathrm{EHVI}>0$, accept; else accept with probability $\exp\bigl(\Delta\mathrm{EHVI}/T\bigr)$.
  \item \textbf{Cooling:} update temperature 
    \[
      T \leftarrow \alpha\,T,
      \quad \alpha = 0.9999,
      \quad T_{0}=1.0.
    \]
\end{enumerate}
We ran $N_{\rm iter}=10^5$ iterations, enforcing uniqueness within each batch at every step.

\paragraph{\textbf{Parallel SA implementation}}

To leverage GPU acceleration, we maintain $M=10$ parallel chains:
\begin{enumerate}
  \item \textbf{Vectorized proposals:} each chain generates $r=2$ perturbed batches in parallel (same perturbation logic as above).
  \item \textbf{Batched EHVI:} all $M\times r$ batches are evaluated in one vectorized qEHVI call on GPU, exploiting batched linear‐operator routines.
  \item \textbf{Independent acceptance:} for each chain, choose its best proposal and apply the same Metropolis criterion.
  \item \textbf{Synchronized cooling:} all chains’ temperatures are multiplied by $\alpha$ each iteration.
\end{enumerate}

\paragraph{\textbf{Hyperparameters and performance}}

\begin{itemize}
  \item Initial temperature $T_{0}=1.0$, cooling rate $\alpha=0.9999$.
  \item Number of chains $M=10$, proposals per chain $r=2$.
  \item Perturbation sizes $m\in\{1,2,3\}$ with $P(m)=[0.6,0.3,0.1]$.
  \item Total iterations: $N_{\rm iter}=10^5$ for sequential; $N_{\rm iter}=2\times10^4$ for parallel (fewer due to GPU speed).
\end{itemize}

\begin{figure}[H]
  \centering
  \includegraphics[width=0.8\textwidth]{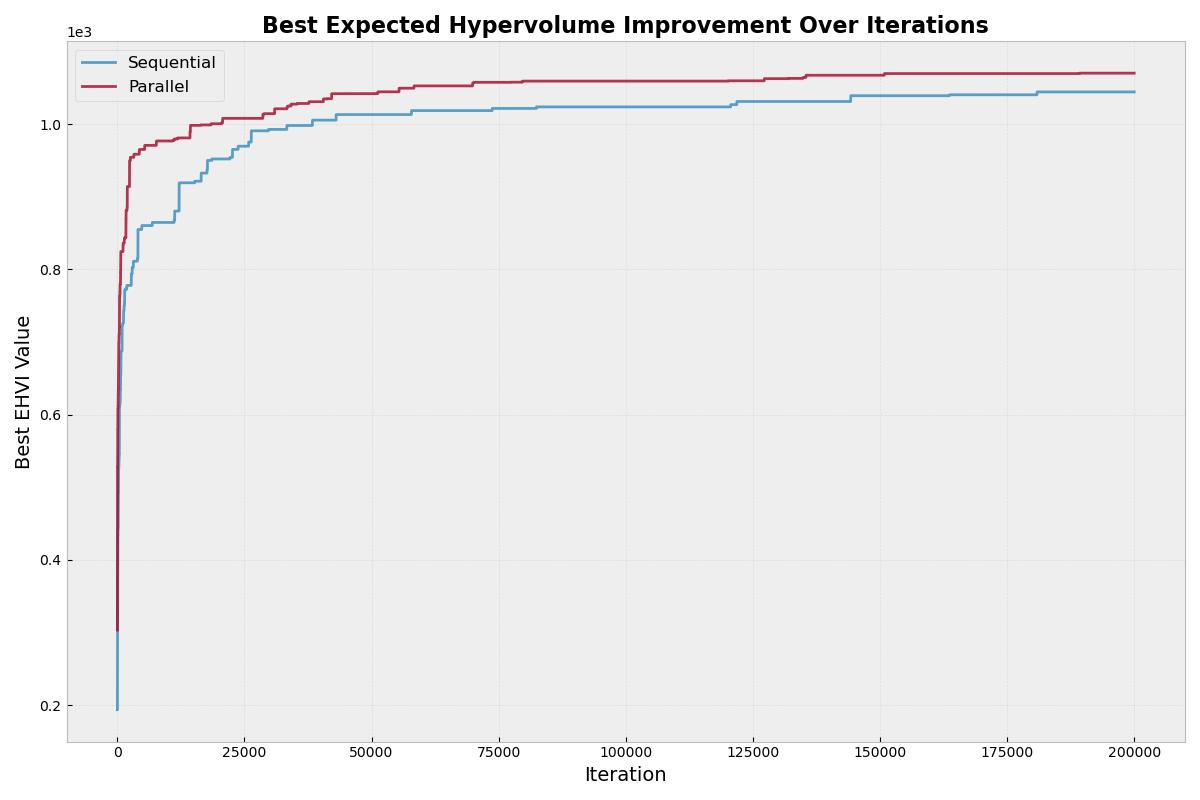}
  \caption{Comparison of sequential versus parallel simulated annealing implementations for a real-world 5-objective materials optimization campaign. The parallel implementation (red) demonstrates superior convergence characteristics compared to the sequential approach (blue), achieving higher final hypervolume values and faster convergence through GPU-accelerated candidate evaluation.}
  \label{fig:parallel_sequential}
\end{figure}

The results presented in Figure~\ref{fig:parallel_sequential} demonstrate the substantial advantages of the parallel implementation over the sequential approach. The parallel simulated annealing achieves a final expected hypervolume improvement of approximately 1.075, compared to 1.055 for the sequential implementation. More importantly, the parallel approach demonstrates superior convergence characteristics, reaching high-quality solutions significantly faster than the sequential method.

The parallel implementation's advantages become particularly pronounced in the early stages of optimization, where rapid exploration of the search space is crucial. The GPU-accelerated evaluation of candidate batches enables the algorithm to process significantly more potential solutions per unit time, leading to more efficient exploration of the complex 9-dimensional search space. This enhanced exploration capability translates directly into improved final performance, as evidenced by the consistently higher hypervolume values achieved throughout the optimization process.

The computational efficiency gains of the parallel implementation are especially significant for practical experimental campaigns, where time constraints and resource limitations often restrict the number of optimization iterations that can be performed. By leveraging GPU acceleration, our parallel simulated annealing approach enables researchers to achieve superior optimization results within the same computational budget, or alternatively, to reach equivalent performance levels with significantly reduced computation time.

These results validate the practical effectiveness of our simulated annealing approach for real-world multi-objective optimization applications, particularly those involving large batch sizes and high-dimensional search spaces. The ability to scale efficiently through parallelization while maintaining strict adherence to discrete candidate sets makes our approach particularly well-suited for experimental optimization campaigns in materials science and related fields where discrete design choices and parallel evaluation capabilities are essential.

%% file: 04_conclusions.tex
\section{Conclusion}
\label{sec:conclusion}
In this work, we have introduced a simulated annealing (SA)–based framework for candidate optimization in batch acquisition functions, addressing key limitations of traditional gradient‐based methods such as SLSQP.  By operating directly on discrete candidate sets and leveraging both sequential and GPU‐accelerated parallel SA implementations, our approach maintains strict adherence to experimental constraints while providing robust global search capabilities.  Across four challenging multi‐objective benchmarks (ZDT1, DTLZ2, Kursawe, and Latent‐Aware), SA consistently matched or outperformed SLSQP, with particularly pronounced gains on high‐dimensional and highly nonconvex landscapes.  Furthermore, in a real‐world five‐objective materials‐design campaign, the parallel SA demonstrated faster convergence and higher final hypervolume, achieving a near 2\% improvement over the sequential version within a fraction of the wall‐clock time.

These results underscore the promise of meta-heuristic optimization for batch Bayesian optimization, especially when large batch sizes or discrete candidate constraints render gradient information unreliable or unavailable.  Future work will explore adaptive cooling schedules, hybridization with local search heuristics, and integration with alternative acquisition functions (e.g., qNoisyEHVI) to further accelerate convergence.  We also plan to extend our parallel SA engine to distributed computing environments and to develop open‐source tooling for seamless integration into existing Bayesian optimization libraries.  Overall, our findings demonstrate that simulated annealing offers a practical, scalable, and effective alternative for multi‐objective batch acquisition optimization in both synthetic benchmarks and real‐world experimental campaigns.